# Ammonia-Net: A Multi-task Joint Learning Model for Multi-class Segmentation and Classification in Tooth-marked Tongue Diagnosis

Shunkai Shi[1,2], Yuqi Wang[3], Qihui Ye[3], Yanran Wang[4], Yiming Zhu[1], Muhammad Hassan[5], Aikaterini Melliou[3], Dongmei Yu[1*]

[1]School of Mechanical, Electrical & Information Engineering, Shandong University

[2]SDU-ANU Joint Science College, Shandong University

[3]Institute of Biopharmaceutical and Health Engineering, Tsinghua Shenzhen International Graduate School, Tsinghua University

[4]School of Software Engineering, Electronic and Information Engineering, Xi'an Jiaotong University

[5]Department of Radiology, Shenzhen Children's Hospital

## Highlights

Traditional Chinese medicine tongue diagnosis can aid in the detection of visceral diseases. However, as the diagnostic process relies heavily on a practitioner's experience, the conventional manual approach may lack reliability.

We propose a novel multi-task joint learning model featuring multi-class segmentation and classification capabilities, along with the ability for tasks to mutually benefit from one another.

To the best of our knowledge, this is the first attempt to apply the semantic segmentation results of tooth marks for tooth-marked tongue classification.

We build a novel tooth-marked tongue diagnosis dataset including 856 images of tongue collected from 856 subjects. The proposed model, namely Ammonia-Net, trains on the acquired dataset together with a comparison study of competitor models.

## Abstract

In Traditional Chinese Medicine, the tooth marks on the tongue, stemming from prolonged dental pressure, serve as a crucial indicator for assessing qi (yang) deficiency, which is intrinsically linked to visceral health. Manual diagnosis of tooth-marked tongue solely relies on physicians' experience. Nonetheless, the diversity in shape, color, and type of tooth marks poses a challenge to diagnostic accuracy and consistency. To address these problems, herein we propose a multi-task joint learning model named Ammonia-Net. This model employs a convolutional neural network-based architecture, specifically designed for multi-class segmentation and classification of tongue images. Ammonia-Net performs semantic segmentation of tongue images to identify tongue and tooth marks. With the assistance of segmentation output, it classifies the images into the desired number of classes: healthy tongue, light tongue, moderate tongue, and severe tongue. As far as we know, this is the first attempt to apply the semantic segmentation results of tooth marks for tooth-marked tongue classification. To train Ammonia-Net, we collect 856 tongue images from 856 subjects. After a number of extensive experiments, the experimental results show that the proposed model achieves 99.06% accuracy in the two-class classification task of tooth-marked tongue identification and 80.02% accuracy in the four-class classification task of tooth-marked tongue ranking, where the macroscopic mean AUC value reaches 0.921. As for the segmentation task, mIoU for tongue and tooth marks amounts to 71.65%.

**Keywords**

Multi-task joint learning; Tooth-marked tongue diagnosis; Multi-class semantic segmentation; Multi-class classification

1. **Introduction**

For millennia, Traditional Chinese Medicine (TCM) theories have been extensively applied in the diagnosis of diseases and serve as a complementary and alternative methodology in conjunction with modern medical practices[1]. The World Health Organization (WHO) has included TCM in its global medical outline (Ver. 2019)[1-3].

Tongue diagnosis [4] is one of the most valuable diagnostic methods in TCM, with tooth marks being considered important indicators of qi (yang) deficiency that reflect the well-being of the internal organs [5, 6]. Clinical professionals can quickly understand a patient's visceral status by simply examining their tongue. As a non-invasive method, tongue diagnosis is rapid, cost-effective, painless, and injury-free, helping to screen various diseases and detect internal organ issues promptly, which is of great value to human health [1, 3, 6]. Accurate diagnosis of the tooth-marked tongue (TMT) is crucial for effective treatment. However, the traditional manual method has the challenges of (1) subjectivity: TMT diagnosis relies heavily on the physician's knowledge and experience which is prone to errors [7]; (2) limitations: the diagnostic experience and knowledge cannot be preserved quantitatively [8], and it is difficult to transfer and train specialized physicians in a short time; (3) instability: external conditions, such as changes in light and angle of vision, can affect the physician's diagnosis [6]. The problems mentioned above lead to issues with consistency and accuracy in the clinical diagnosis of tooth-marked tongue [1]. Addressing the subjective and empirical dependence in the diagnosis of the tooth-marked tongue would facilitate the broader use of tongue diagnosis in modern medicine worldwide [9]. An automated and reliable diagnostic approach is essential to meet clinical needs.

Artificial intelligence and computer vision have developed rapidly and been widely used for medical information analysis in the past decade to ascertain information within a dataset [10-17]. There are extensive studies have been carried out in this line of work. For example, semantic segmentation techniques are used to identify lung lesions caused by COVID-19 [18], and classification models are used to analyze images of infectious keratitis [19]. Deep learning is extensively applied in the field of tongue diagnosis as well, including tooth-marked tongue diagnosis [1, 20-23], tongue color classification [24-26], and joint diagnosis of multiple conditions [27-32]. However, the application of computer vision in the tooth-marked tongue diagnosis has the following challenges: (1) the tooth marks along the tongue's edge are relatively small in comparison to the tongue's overall size; (2) there is a tiny color difference between the tooth marks and the surroundings; (3) challenging to distinguish the tooth-marked tongues of different classes susceptible to boundaries; (4) no standard illumination and orientation while capturing images, thus large differences exist between images, resulting in diverse representations of the same features.

Fig. 1 illustrates the subjects' tongue images and the division into four classes based on the tooth marks on the tongue. The HT, LT, MT, and ST indicate healthy tongues without tooth marks, and light, moderate, and severe tooth-marked tongues, respectively.

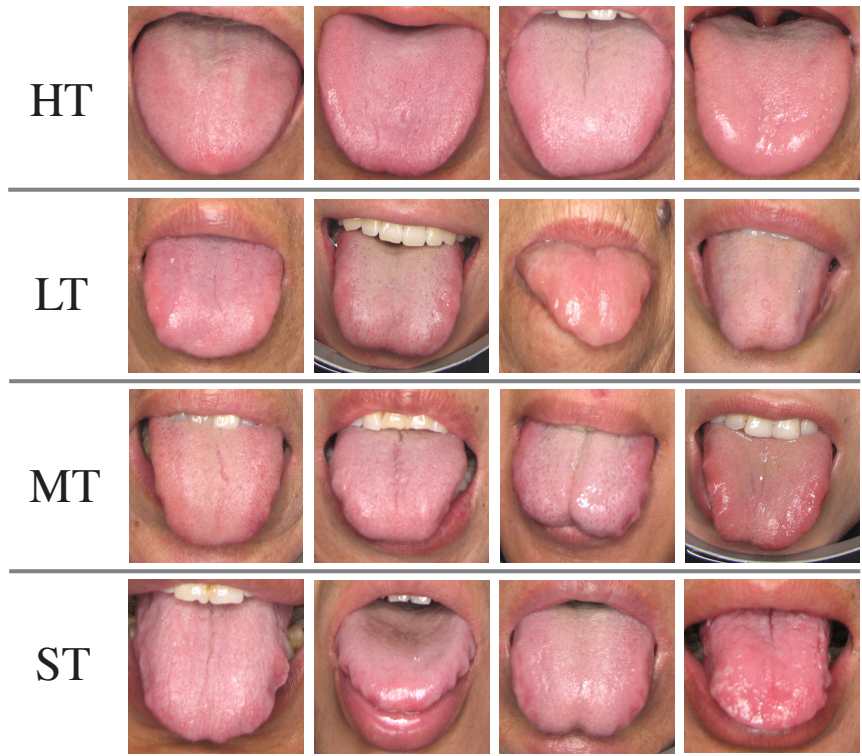

**Fig. 1. Representative RGB tongue images captured using standard cameras. The four classes are Healthy Tongue(HT), Light Tongue(LT), Moderate Tongue(MT), and Severe Tongue(ST).**

We propose a multi-task joint learning model, namely Ammonia-Net resembling $NH_3$ as the input involves three coordinated processes, which allows multi-class semantic segmentation and classification of tongues with tooth marks. The workflow of Ammonia-Net follows as: first, the input image is learned by the segmentation branch and the binary classification branch, after which the original image is concatenated with the output feature map of the segmentation task in the channel dimension to obtain a six-dimensional tensor. The resulting six channels are derived from both the RGB channels of the initial image and the three channels represent different features within the output feature map. The resulting tensor with a channel dimension of six is learned by a four-class classification block. Second, the output of the two-class module is fused with the output of the four-class module to obtain the enhanced four-class outcome. Finally, Ammonia-Net modifies the segmentation output according to the class information of the corresponding images given in the classification section.

The key contributions of our work are articulated as follow:

1. We propose a multitask joint learning model, namely Ammonia-Net, for TCM tooth-marked tongue classification as well as tooth marks and tongue segmentation using RGB tongue images, which enables combining semantic segmentation and classification tasks to improve the effectiveness of both segmentation and classification.

2. We introduce a customized multi-task loss function and outperforms the models trained with traditional objective functions.

To the best of our knowledge, the proposed model attempts for the first time to apply the semantic segmentation results of tooth marks for tooth-marked tongue classification

3. We build a novel tongue image dataset consisting of 856 images collected from 856 subjects to demonstrate the superiority of the Ammonia-Net over other state-of-the-art methods in comprehensive multi-metric comparison experiments.

## 2. Related Works

This study is based on the multi-task joint deep learning of semantic segmentation and classification for diagnosis of tooth-marked tongue using images. The related work to TMT divides into manual diagnosis, automated methods, and multi-task learning.

2. 1 Manual diagnosis of tooth-marked tongue

Tongue diagnosis is one of the most meaningful and important diagnostic methods in TCM, and its results of the TMT are considered to help clinicians understand the status of various organs in a rapid and non-invasive manner. For thousands of years, it has been widely used in clinical analysis of TCM [4, 33, 34]. More specifically, the tooth-marked tongue is an important indicator in TCM diagnosis when assessing qi (yang) deficiency, especially in spleen-qi (yang) deficiency, which is considered to be intrinsically related to dysfunction of the digestive system [5, 6]. In addition, symptoms of tooth-marked tongue consist of tissue edema, inadequate blood supply, and local hypoxia. Loss of appetite, toothache, and other clinical manifestations commonly seen in patients with tooth-marked tongue [22]. A recent study has shown that tongue diagnosis can even assist in cancer screening [35].

Based on the location, visual depression level, and the number of tooth marks, the tooth-marked tongue is divided into three classes by TCM practitioners, and each class has a unique guiding significance for the diagnosis of visceral diseases [6]. The traditional manual diagnosis of tooth-marked tongue relies heavily on physician experience [36]. And the illumination and orientation while observing easily interfere with the diagnosis result in inconsistent and unreliable results [1].

2. 2 Automated methods for tongue diagnosis

Traditional algorithms is used to obtain tongue contours, locate toothmarks, and identify tooth-marked tongue [6, 21, 37, 38], but their performance is susceptible to interference from factors such as light. Li et al [36] use indentation to identify suspicious regions, extract feature vectors using ConvNet, and then implement tooth marks recognition and tooth-marked tongue identification using Multiple Instance Learning. This method improves the robustness but requires human selection of pixel thresholds thus less automation. Kong et al [39] implement automatic segmentation of tooth marks based on Mask Scoring R-CNN and migration learning, but tooth marks are not used for the classification of the tooth-marked tongue. Tang et al [1] perform cascaded CNN and multi-task learning for the tongue as well as tooth marks detection. The results assist tooth-marked tongue recognition without accurate segmentation of tooth marks. Wang et al [22] implement tooth-marked tongue determination using ResNet34 and demonstrated CNN focus on tooth marks with Grad-CAM. However, the tongue needs to be manually segmented in some method [22, 36-38], and accurate tooth marks information is not used for tooth-marked tongue classification.

Traditional algorithms for tongue segmentation include active contour model-based methods [40, 41], edge-based methods [42, 43], region-based methods [44], and mixed methods [45]. Lin et al [46] design DeepTongue based on ResNet and achieve end-to-end tongue segmentation without preprocessing compared to traditional algorithms. Xue et al [44] use DeepLabV3 and perform accurate segmentation. CNN that enables fast localization and then segmentation of tongues is proposed [47, 48]. To improve the segmentation efficiency, Cai et al [49] propose a new loss function involving inter-class and intra-class costs to replace the cross-entropy loss function. It is proven to be easy to integrate into various state-of-the-art networks. Huang et al [50] design a new residual soft-link module based on U-Net to reduce the semantic differences in the feature map fusion process. Then the significant image fusion module is applied to preserve more image details. Further, the new loss function is used to enhance the performance of segmenting different resolution tongues. To improve practicality, Qiu et al [51] design a feature extraction framework combining classification and segmentation for mobile devices. However, these methods do not apply fine tooth marks to the tooth-marked tongue classification and do not correct the impact of tooth marks on tongue segmentation results.

2. 3 Multi-task learning

Multitask learning [52], a machine learning method that uses knowledge from multiple tasks to assist each task[53], is proven effective in improving performance with small data volumes and combining different tasks[54-59]. Multitasking can be implemented in the following ways [60-63]. Specifically, Thrun and Sullivan et al [60] report a method to achieve multitask learning by sharing weights. The feature maps of the corresponding layers from different task networks are combined linearly by channel [61]. Regularization is used for multitask learning to balance different tasks [62]. Yang et al [63] improve performance by sharing features across multiple tasks based on tensor factorization. The multitasking model can be further optimized with strategies for automatically adjusting the shared parameters [64], weighing the loss function [65], and dynamically adjusting the gradient. To reduce the complexity, PackNet [66] packs multiple tasks into the same network by iterative pruning. In computer vision, the multi-task attention framework of Liu et al [67] allows for extracting task-specific messages from global information. Dai et al [68] report a division of instance-aware semantic segmentation into classification and semantic segmentation multi-task cascade networks. Classification in the model proposed by Li et al [69] relies on the prediction of object detection and segmentation, giving a fuller performance of multitasking. In the field of medical diagnosis, multitasking multichannel learning [70, 71] is used for joint classification and score regression of brain MRI. Le et al [72] perform both classification and pixel-level segmentation of mammograms. Amyar et al [73] classify COVID and segment the lesions using CT images of the lungs. The MMCL-Net proposed by Hong et al [74] enables simultaneous object detection, segmentation, and classification of different structures in spine MRI. Xu et al [75] apply tongue segmentation predictions for classification but do not segment detailed features for further multitasking performance.

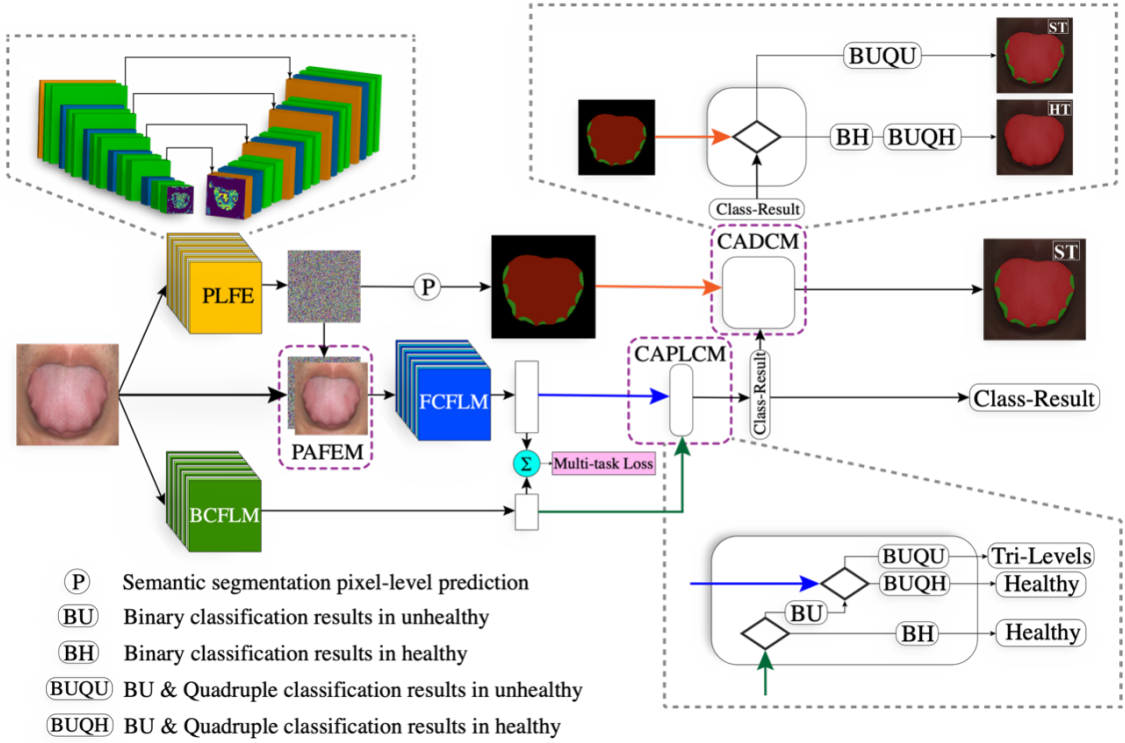

**Fig. 2. Detail diagram of proposed multi-task joint learning model. It mainly includes six modules: Pixel-Level Feature Extractor (PLFE), a U-net, which is used to extract the tongue and tooth marks from RGB tongue images at the pixel level; Pixel Aware Feature Enhancement Module (PAFEM), which is used to concatenate the PLFE output feature maps with the original RGB image in the channel dimensional to highlight the tongue and tooth marks; Binary Classification Feature Learning Module (BCFLM), which is the ShufflenetV2 that classifies the tongue into two classes, namely HT and UH; Quadruple classification feature learning module (QCFLM), which is a modified ShufflenetV2 used to accurately classify the tongue in four classes, namely HT, LT, MT and ST; Class-aware diagnostic correction module (CADCM), which combines the results from BCFLM and QCFLM to improve the classification results; Class-aware pixel-level correction module (CAPLCM), which is used to correct the segmentation results of PLFE at the pixel level.**

## 3. Method

3.1 Architecture of Ammonia-Net

Fig. 2 illustrates the general framework of the proposed multi-task joint learning model (Ammonia-Net) for segmentation and classification. The following are the main modules of the proposed model:

**Pixel-Level Feature Extractor** (PLFE): Heightened attention can be applied to specified features in the original image via U-Net [76], a network model for image segmentation tasks with a symmetric U-shaped structure that includes an encoder to capture contextual information and a decoder to recover spatial details. The high attention will be reflected in the output feature map of U-Net. The feature maps derived after upsampling are valuable for the attention area indication of the classification task while tooth marks have inconspicuous colors. Therefore, we use U-Net as a pixel-aware semantic segmentation module for tongue body and tooth marks. More features are provided for the classification task while accessing segmentation results. In the implementation, we impose three imaging classes, namely tongue, tooth marks, and background.

Pixel Aware Feature Enhancement Module (PAFEM): The purpose of the PAFEM module is to give more distinctive features to the QCFLM module based on the RGB to induce a parameter updating process by highlighting the tooth marks. The number and severity of tooth marks on the tongue vary greatly between classes. In other words, tooth marks are an important feature for classifying the TMT. However, in tongue images, tooth marks are similar in color to the tongue. To increase the focus of the QCFLM module on the tongue and tooth marks, we concatenate the raw RGB images of the tongue with the channel dimensions of the PLFE output to obtain a 6-channel tensor, which is fed to QCFLM. These six channels are derived from the three channels of RGB from the original image, and the three channels represent different features in the U-Net output feature map.

Binary Classification Feature Learning Module (BCFLM): The BCFLM module learns from healthy and unhealthy tongue images. We employ the efficient ShuffleNetV2 [77], a lightweight CNN with remarkable speed and accuracy trade-offs featuring channel shuffling and bottleneck design, to predict the healthiness of the tongue.

Quadruple classification feature learning module (QCFLM): The QCFLM module is a modified version of ShuffleNetV2, which aims at the accurate classification of the tooth-marked tongue. During the learning process, a 6-channel tensor from the PAFEM module serves as the input of the QCFLM. In the implementation, we consider four classes, HealthyTongue (HT), LightTongue (LT), ModerateTongue (MT), and SevereTongue (ST).

Class-aware diagnostic correction module (CADCM): Compared with the quadratic classification task, ShuffleNetV2 is experimentally shown to make more accurate predictions of healthy tongue in binary classification. To make the classification results more reliable, we integrate the outputs of both QCFLM and BCFLM. And we design CAPLCM to achieve the goal of modifying the results of the four-class classification. The module is executed as follows: the output of the BCFLM feeds to the multi-branch pipeline and all outputs about healthy images are recognized, even if the QCFLM gives unhealthy prediction results; the input about unhealthy forwards to the next multibranch pipeline together with the output of the QCFLM, if the QCFLM gives a healthy judgment, we give a healthy prediction for this tongue; otherwise, we give this tongue the a prediction result as one among the three classes of tooth-marked tongues from QCFLM.

Class-aware pixel-level correction module (CAPLCM): Due to the similar color between tooth marks and the tongue, coupled with the absence of spatial cues in RGB imagery, segmenting tooth marks presents a formidable task for the segmentation neural network. Medically, tooth marks are absent from a tongue in a healthy tongue, however, the PLFE is incapable of discerning the wellness of the tongue. Due to illumination, orientation, and other physiological issues, some areas on the tongue can be predicted incorrectly as tooth marks. We use the CADCM module to correct these mistaken tooth marks on healthy tongues, and CADCM performs as follows: CADCM receives prediction results from both the classification and segmentation. Notably, according to the experimental verification, the prediction recall of the classification part about healthy tongue reached 99.3%, so the prediction about healthy tongue is supposed to be always correct. If the classification result shows that the prediction of an image is healthy (HT), then the prediction about tooth marks in the segmentation task will be corrected to the tongue body. High-accuracy prediction results about health will reduce the probability that a healthy tongue will be misdiagnosed in segmentation task. In other words, within the prediction accuracy of the classification network, tongues without tooth marks will not be wrongly diagnosed.

Ammonia-Net works as follows: the resized 512×512-pixel image is the input to the whole network. The segmentation results of tooth marks, tongue body, and a four-dimensional vector representing the prediction of class probabilities are the final output. More specifically, first PLFE performs pixel-level prediction of tooth marks and tongue while obtaining feature maps. The 512 × 512-pixel image is also input to BCFLM for classification learning based on two labels: unhealthy tongue (UH) and healthy tongue (HT), after which PAFEM performs feature enhancement operations on the input RGB tongue image using the outcome feature map of the segmentation part to generate pixel-level enhancement data RGBWPF. RGBWPF serves as the input of the subsequent QCFLM, which learns classification based on four labels (HT, LT, MT, ST) for RGBWPF. Notably, the losses of BCFLM and QCFLM jointly constitute a multitask loss. The classification outputs of BCFLM and QCFLM are integrated by CAPLCM to give more accurate results for the four classifications. The prediction of CAPLCM at the class level is fed to CADCM for correction on the segmentation outcomes.

3. 2 Loss function

The Ammonia-Net network tunes the parameter via objective function is formulated as follows:

$$L_{MTL} = \alpha L_B + \beta L_Q \tag{1}$$

Where, $\alpha, \beta$ are the weights, and both are taken as 1 according to experimental verification.

The cumulative loss is composed of two loss terms including $L_B$ and $L_Q$. The $L_B$ is obtained in the two-class classification task by calculating the cross-entropy loss of the prediction results with ground truth. And $L_Q$ is obtained in the same way in four-class classification. $L_{MTL}$ is generated by a linear combination of $L_B$ and $L_Q$. In addition The PLFE is first trained separately and reloaded to obtain weights to the PLFE. When optimizing the model based on $L_{MTL}$, the parameters of the PLFE are frozen. In other words, the $L_{MTL}$ only works for the parameter updates of BCFLM and FCFLM. This approach also helps in gradient descent optimization, convergence in multi-task learning, and sustaining the training speed of lightweight classification.

## 4. Experiment

### 4.1 Dataset

In collaboration with the Shanghai University of Traditional Chinese Medicine, we collected a tongue image dataset Tooth-marked Tongue Diagnosis 2020 (TTD-2020) for tooth-marked tongue classification and tongue body as well as tooth marks semantic segmentation. These images are RGB images captured using standard cameras. All images are annotated by three medical professionals. The masks used for segmentation are tagged with Labelme for the tongue body and tooth marks. As shown in Table 1, the TTD-2020 dataset consists of four classes: healthy tongue (HT), light tongue (LT), moderate tongue (MT), and severe tongue (ST). A total of 856 images are derived from 856 subjects.

Table 1 Details of the TTD-2020 dataset

| Class name | Number of subjects | Number of images |
|:---:|:---:|:---:|
| HealthyTongue (HT) | 291 | 291 |
| LightTongue (LT) | 133 | 133 |
| ModerateTongue (MT) | 233 | 233 |
| SevereTongue (ST) | 199 | 199 |
| Total | 856 | 856 |

### 4.2 Evaluation metrics

To evaluate the performance of the underlying segmentation models, we employ IoU, Recall, Precision, and $F_\beta$-score. Similarly, for classification, we use four metrics, which are: (1) Accuracy; (3) Recall; (2) Precision; (4) $F_\beta$-score.

In the following equations TP, FN, FP, and TN represent True Positive, False Negative, False Positive, and True Negative respectively.

Intersection over Union (IoU) serves as a common metric for assessing performance in image semantic segmentation tasks, which is defined as:

$$IoU = \frac{TP}{FP + TP + FN} \tag{2}$$

Accuracy refers to the percentage of successfully classified images to the total number of images, and the formula is:

$$Accuracy = \frac{\sum_{i=0}^{L-1} x_{ii}}{\sum_{i=0}^{L-1} \sum_{j=0}^{L-1} x_{ij}} \tag{3}$$

$x_{ij}$ refers the $(i,j)$-th value in the confusion matrix and $L$ stands for the number of classes.

The recall is defined as follows and refers to the average recall for each class.

$$Recall = \frac{1}{L} \sum_{0}^{L-1} \frac{TP_l}{TP_l + FN_l} \tag{4}$$

Where $TP_l$ is the false positive rate about the class $l$.

Precision is the average precision of each class, defined as:

$$Precision = \frac{1}{L} \sum_{l=0}^{L-1} \frac{TP_l}{TP_l + FP_l} \tag{5}$$

$FP_l$ in formula (5) is the number of false positive about the class $l$.

$F_\beta$-score is the combined value of recall and precision, defined as:

$$F_\beta - score = (1 + \beta^2) \times \frac{Precision \times Recall}{\beta^2 \times Precision + Recall} \tag{6}$$

Where the value of $\beta$ represents the degree of attention to precision and recall, and the larger $\beta$ means more attention to recall. In this task, we take $\beta = 1$ in order that the $F_\beta$-score can reflect the overall performance of recall and precision, i.e., we use $F_1$-score.

## 4.3 Implementation details

### 4.3.1 Data Augmentation

In order to reduce the risk of overfitting and improve the model, we use augmentation methods for both semantic segmentation and classification tasks. The augmentation strategies used in the classification task are (1) random horizontal flipping; (2) random cropping and scaling to a size of 512 × 512-pixel. For the segmentation, both ground truth and mask versions are passed for the augmentation as (1) the scaling and aspect warping in the range of (0.25, 2); (2) flipping left and right with 75% probability and flipping up and down with 75% probability; (3) randomly rotating in the range of (-180°, 180°) with 90% probability; (4) performing one of the four drift operations of brightness, saturation, contrast, and hue in the range of 0.1 with 67% probability of equal chance.

### 4.3.2 Training Configuration

All the underlying models are implemented with the Nvidia RTX 2080Ti GPU with Pytorch. All images are resized to 512 × 512-pixel before input to the model. During the segmentation model, i.e. U-Net training, the Adam optimizer tunes the parameters of the exponential weighted moving average = 0.9, the decay rate for the exponential weighted moving average of squared gradients = 0.999, and a learning rate = 0.1. In addition, we use cosine annealing to reduce noise perturbations which helps the model in fast convergence and maintains the generalizability. In order to increase the modularity and flexibility of the multi-task framework and improve the accuracy of the gradient descent direction in the optimization for the classification model, we divide the training process into two phases: first, optimizing PLFE for the segmentation task of tooth marks and the tongue segmentation for 1000 epochs. The second phase is the training of the classification network, in which we import the parameters of the trained U-Net into PLFE and freeze them. It means that the parameters of PLFE are no longer changed. To update the parameters of the classification part, we use SGD optimizer with initial learning rate = 0.006 and weight decay factor = $4 \times 10^{-5}$ deploying 1000 epochs.

4. 4 Comparison study with state-of-the-art methods

To verify the effectiveness of the proposed method in classification and segmentation. We train several state-of-the-art (SOTA) classification and segmentation networks for comparison, including MobileNetV2 [78], RegNet [79], SwinTransformer-Tiny [80], AlexNet [81], Efficientnet [82], DenseNet121 [83], ResNet34 [84], ShuffleNetV2 [77]; Unet++ [85], Deeplabv3+ [86], $U^2$-Net [87], U-Net [76]. For all these models, trainings and tests are performed according to the default settings, with the same input image size of 512 × 512-pixel. In addition, all models are validated using a ten-fold cross-validation, with the data set divided in the same way as the experiments of the proposed model.

4. 4. 1 Results of four-class classification

The quantitative results for the quadratic classification are shown in Table 2. The model proposed outperforms the competitor's models in all four metrics. Ammonia-Net achieves an accuracy of 80.02% and is the only approach that exceeds 80.00% of all methods, which is a 4.63% improvement over the second-best method, ShuffleNetV2. In terms of precision, Ammonia-Net obtained 79.54%, an improvement of 8.69% over the ShuffleNetV2, the second-best method. Notably, the recall score of the proposed method is 10.24% higher than the second-best method EfficientNetb6, reaching 76.87%. In terms of the $F_1$-score, the proposed method obtained 76.85%, an improvement of 9.19% over the second-best method ShuffleNetV2, making it the only model with an $F_1$-score above 75.00%.

**Table 2 Comparison of the performance achieved by the model proposed with the state-of-the-art methods in the test set of the TTD-2020 dataset in Quad-Classification. Bold black text indicates the highest scores of all methods in the corresponding metrics.**

| Method | Accuracy | Precision | Recall | $F_1$-score |
| --- | --- | --- | --- | --- |
| Mobilenet-V2 | 0.3971 | 0.2597 | 0.2425 | 0.3289 |
| Regnet | 0.4042 | 0.2488 | 0.2519 | 0.3328 |
| SwinTransformer-Tiny | 0.4906 | 0.5065 | 0.4610 | 0.4811 |
| Alexnet | 0.5129 | 0.5089 | 0.4574 | 0.4999 |
| Efficientnetb7 | 0.5977 | 0.6071 | 0.5747 | 0.5780 |
| Efficientnetb0 | 0.6189 | 0.5892 | 0.5181 | 0.5437 |
| Efficientnetb2 | 0.6305 | 0.6461 | 0.5929 | 0.5895 |
| Efficientnetb5 | 0.6664 | 0.6488 | 0.6312 | 0.6394 |
| Efficientnetb4 | 0.6680 | 0.6300 | 0.5711 | 0.5853 |
| Densenet121 | 0.6711 | 0.6317 | 0.5937 | 0.6011 |
| Efficientnetb1 | 0.6778 | 0.6661 | 0.6263 | 0.6234 |
| Resnet34 | 0.6855 | 0.6767 | 0.6332 | 0.6421 |
| Efficientnetb3 | 0.6900 | 0.6641 | 0.6410 | 0.6419 |
| Efficientnetb6 | 0.7040 | 0.6794 | 0.6663 | 0.6740 |
| ShufflenetV2 | 0.7539 | 0.7085 | 0.6586 | 0.6766 |
| **Ammonia-Net** | **0.8002** | **0.7954** | **0.7687** | **0.7685** |

Fig. 3 shows the normalized confusion matrix obtained by Ammonia-Net after ten-fold cross-validations for the quadratic classification tasks of the tooth-marked tongue. A recall of 99.3% is achieved for HT, indicating that the proposed model can distinguish between control and tooth-marked tongues well. The model also achieves recall of 66.2%, 75.5%, and 66.3% for LT, MT, and ST, respectively. In addition, no UH samples are predicted to be HT, indicating that the proposed model does not miss any patients.

Fig. 4 shows the ROC (receiver operating characteristic) curves on the quadratic classification tasks for the tooth-marked tongue classification. The AUC (area under the curve) value for the HT reaches 0.998, which demonstrates that the Ammonia-Net can discriminate well between healthy and tooth-marked tongues. For the other classes, the proposed model achieves AUC values of 0.889, 0.859, and 0.936 for LT, MT, and ST, respectively. For the macroscopic average AUC value, it reaches 0.921, which indicates that Ammonia-Net outperforms in the four-class classification task of the tooth-marked tongue as shown in Fig. 4.

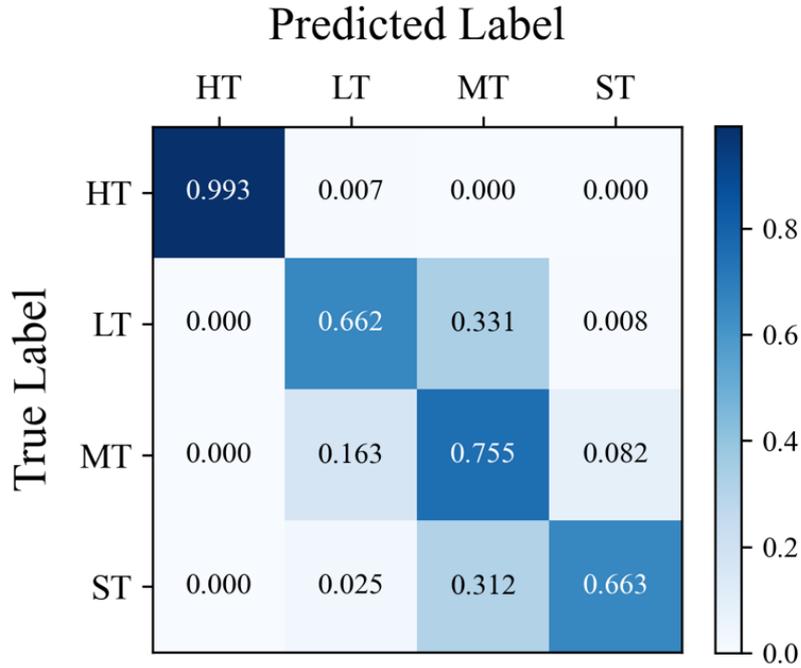

Fig. 3. Confusion matrix obtained by testing Ammonia-Net on the TTD-2020 dataset test set for the four-classification task of tooth-mark tongue classification. The rows represent the true labels tagged by the professionals and the columns represent the predicted results generated by the model proposed. The blue color deepens with increasing values in the matrix

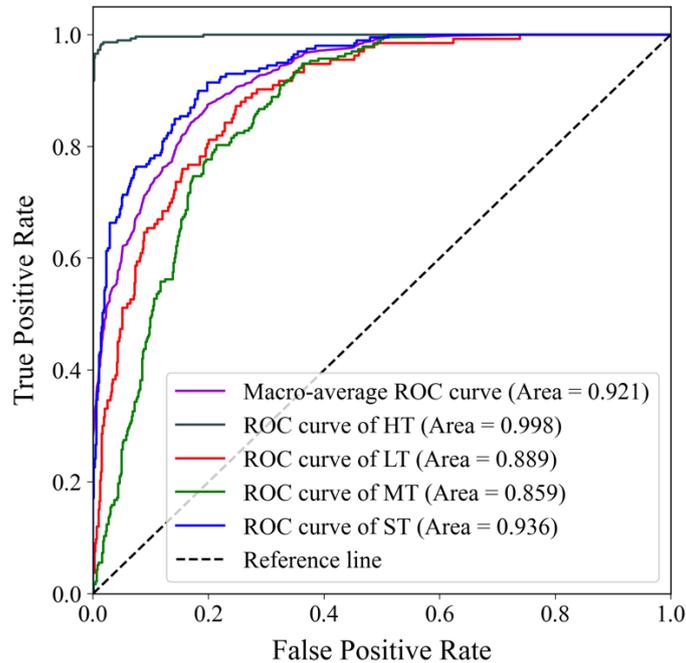

Fig. 4. ROC curves obtained from Ammonia-Net tested on the TTD-2020 dataset test set for the quadratic classification task of the tooth-marked tongue. The Area refers to the area under the corresponding curve.

The qualitative results of the four-class classifications are as follows:

Gradient-weighted Class Activation Mapping (Grad-CAM), a visualization technique for interpreting neural network prediction results. It calculates gradients to generate heat maps that highlight regions highly relevant to the target class, showing whether the model focuses on medically agreed key features. The Grad-CAM generated using other state-of-the-art models to be compared with the method proposed for some representative cases are shown in Fig. 5. Darker-colored (blue) regions indicate pixels receiving less attention, and red more attention. By comparing the tooth marks labeled by medical experts with the Grad-CAM, the real tooth marks are more consistent with the areas of attention demonstrated by the Grad-Cam in our model. This indicates that Ammonia-Net captures medically agreed key features (tooth marks) on a quadratic classification task of tooth marks tongue, which increases the interpretability and feasibility of the proposed model.

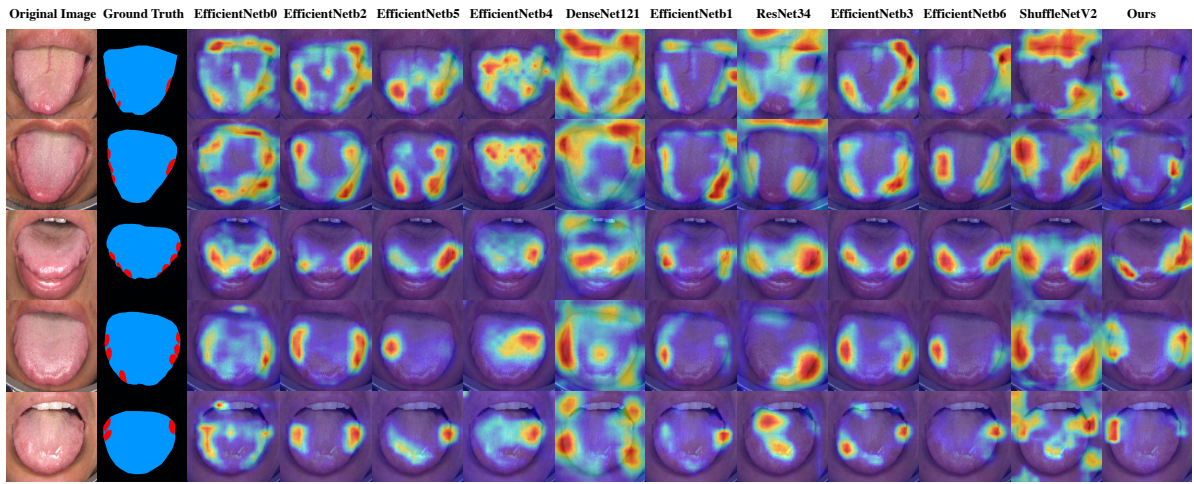

**Fig. 5. CAM, visualization results obtained for different methods. The first column counting from left to right is the original image, and the second column is the tongue(blue) and tooth marks(red) labeled by professional physicians. The remaining images are the CAM heatmaps obtained on representative images of each model.**

4. 4. 2 Results of two-class classification

The quantitative results:

Ammonia-Net obtains quantitative results in the tooth-marked tongue binary (healthy tongue, unhealthy tongue) classification task as shown in Table 3. In ten-fold cross-validation on the TTD-2020 dataset, Ammonia-Net achieves 99.06% accuracy, 99.22% precision, recall reaches 98.80%, and $F_1$-score attains 98.88%.

Fig. 6 shows the normalized confusion matrix obtained after performing ten-fold cross-validation on Ammonia-Net for the tooth-marked tongue binary classification task. And UH achieves a recall of 100%, which indicates the absence of misclassification critical for medical diagnosis tasks. Meanwhile, Ammonia-Net achieves a 97.3% recall score for HT.

Table 3 The scores of our model on the binary classification of HT and UH in terms of four metrics.

| Method | Accuracy | Precision | Recall | $F_1$-score |
|---|---|---|---|---|
| Ammonia-Net | 0.9906 | 0.9922 | 0.9880 | 0.9888 |

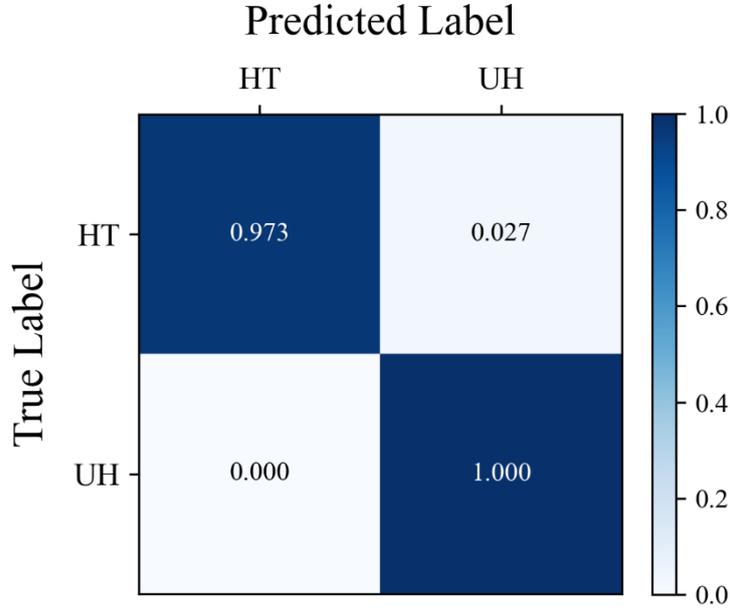

Fig. 6. Confusion matrix obtained by Ammonia-Net tested on the TTD-2020 dataset test set for the binary classification task of the tooth-marked tongue. The rows represent the true labels tagged by the professionals and the columns represent the predicted results generated by the proposed model. The blue color deepens with increasing values in the two-dimensional matrix.

4. 4. 3 Results of the segmentation task

Table 4 Comparison of the performance achieved by the proposed model with the state-of-the-art methods in the test set of the TTD-2020 dataset in Segmentation. Bold black text indicates the highest scores of all methods in the corresponding metrics.

| Method | Classes | $F_1$-score | Precision | Recall | IoU |
|---|---|---|---|---|---|
| U-Net++ |  | 86.13 | 84.73 | 87.57 | 75.63 |
| Deeplabv3+ |  | 95.73 | **98.31** | 93.28 | 91.81 |
| U$^2$-Net | Tongue | 97.73 | 98.03 | 97.44 | 95.53 |
| U-Net |  | 97.73 | 97.79 | 97.67 | 95.56 |
| **Ammonia-Net** |  | **97.77** | 97.72 | **97.81** | **95.63** |
| U-Net++ |  | 22.04 | 13.68 | 56.68 | 12.38 |
| Deeplabv3+ |  | 41.88 | 28.14 | **82.46** | 26.52 |
| U$^2$-Net | ToothMarks | 62.71 | 60.41 | 65.31 | 45.66 |
| U-Net |  | 63.69 | 62.47 | 65.18 | 46.75 |
| **Ammonia-Net** |  | **64.54** | **64.02** | 65.29 | **47.67** |

The quantitative results for the segmentation task are shown in Table 4.

For the class of tooth-marked tongue, the model proposed outperforms the competitor models in terms of $F_1$-score, Precision, and IoU. The proposed method obtains an $F_1$-score score of 64.54%, which is 0.85% better than the second-best method U-Net. For Precision, our method obtains 64.02%, an improvement of 1.55% over the second-best method U-Net. Regarding the toothmark's IoU, our method achieves 47.67%, which is 0.92% better than the second-best method U-Net.

For the class of tongue, Ammonia-Net outperforms competitor models in terms of $F_1$-score, Recall, and IoU. For $F_1$-score, the proposed method achieves 97.77%, which is 0.04% better than the second-best method U-Net. For recall, Ammonia-Net obtains 97.67%, which is 0.14% better than U-Net, the second-best method. For IoU, the proposed method achieves 95.63%, which is 0.07% better than the second-best method U-Net.

The qualitative results are as follows:

Fig. 7 presents a visualization and intuitive comparison of the segmentation results for different methods regarding different classes of tongues on the validation set. It is observed that Ammonia-Net has an advantage over previous state-of-the-art methods in tooth mark segmentation and achieves better results in tongue segmentation.

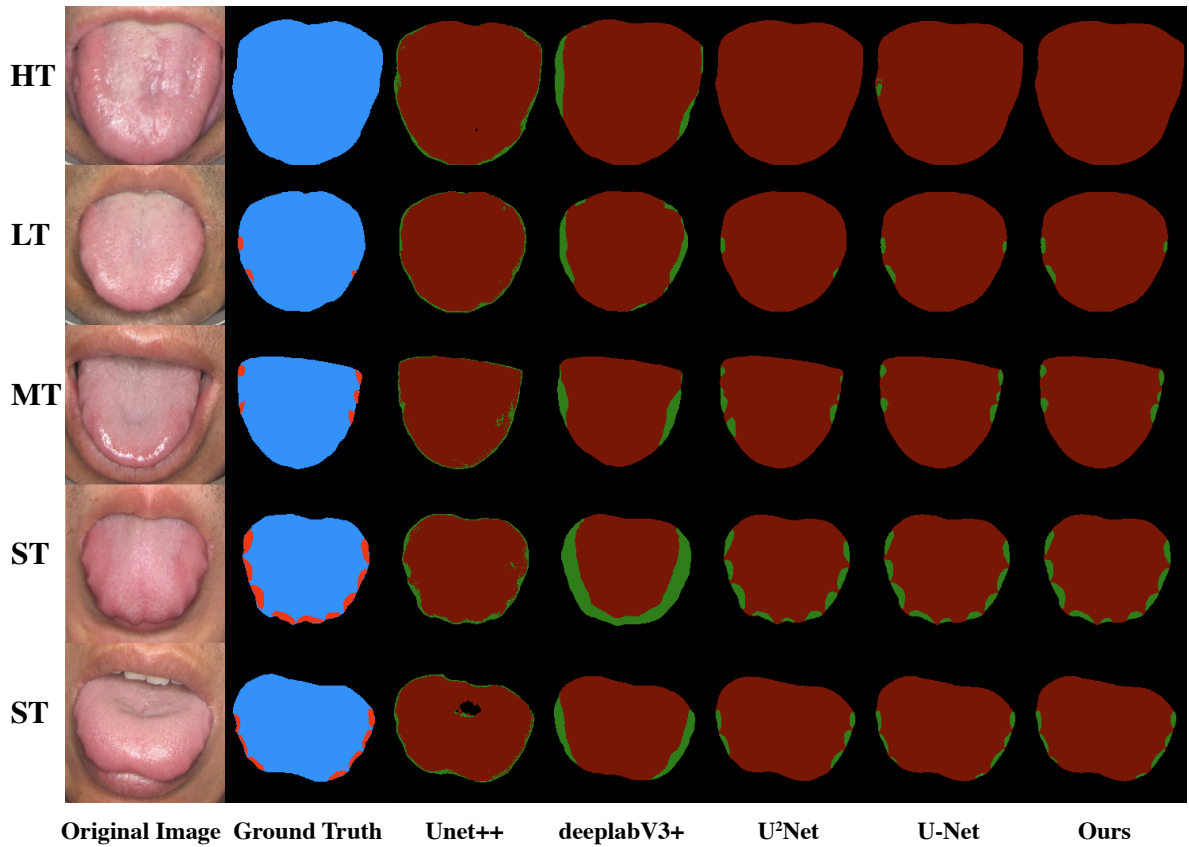

**Fig. 7. Visualization segmentation results for each method on different classes of tongues. Where, from left to right, the first column shows the RGB tongue images and the second column shows the tongue(blue)and tooth mark(red) labeled by specialists. The rest of the images are the visualization segmentation results obtained by different models on different classes of tongues, red represents**

4.5 Ablation experiments

There are five main improvements in Ammonia-Net, which are CADCM, PAFEM, BCFEM, CAPLCM, and multi-task training (MTT). To verify the effectiveness of each module, we compare a total of 15 different networks for classification and segmentation tasks, respectively. Clab (backbone network of classification, which is shufflenetV2) and Segb (backbone network of segmentation, which is U-Net) function as benchmark standards for the ablation experiments, focusing on classification and segmentation tasks individually.

4.5.1 Ablation Results of the classification task

The results of the ablation experiments for the classification task are shown in Table 6.

**Table 6 Presentation details of different models for classification tasks. Bold black text indicates the highest scores of all methods in the corresponding metrics.**

| No | Method | Accuracy | Recall | Precision | $F_1$-score |
|---|---|---|---|---|---|
| 1 | Clab | 0.7539 | 0.6586 | 0.7085 | 0.6766 |
| 2 | Clab+MTT | 0.7443 | 0.7023 | 0.7280 | 0.7008 |
| 3 | Clab+CAPLCM | 0.7603 | 0.6761 | 0.7035 | 0.6484 |
| 4 | Clab+PAFEM | 0.7771 | 0.7269 | 0.7424 | 0.7269 |
| 5 | Clab+PAFEM+MTT | 0.7662 | 0.7437 | 0.7627 | 0.7391 |
| 6 | Clab+CAPLCM+MTT | 0.7684 | 0.7172 | 0.7535 | 0.7156 |
| 7 | Clab+PAFEM+CAPLCM | 0.7986 | 0.7455 | 0.7661 | 0.7378 |
| 8 | Clab+PAFEM+CAPLCM+MTT | **0.8002** | **0.7954** | **0.7687** | **0.7685** |

**MTT effectiveness:** as shown in Table 6, when MTT is added to the base model, by comparing No2 and No1, recall improved by 4.37%, precision improved by 1.95%, and $F_1$-score improved by 2.42%; when MTT is added to Clab+PAFEM, by comparing No5 and No4, recall improved by 1.68%, precision by 2.03%, and $F_1$-score by 1.22%; when MTT is added to Clab+PAFEM+CAPLCM, by comparing No8 and No7, accuracy improved by 0.16%, recall by 4.99%, precision by 0.26%, and $F_1$-score by 3.07%.

**CAPLCM effectiveness:** in the case of Table 6, when CAPLCM is added to the base model, by comparing No2 and No1, accuracy improved by 0.64%, recall improved by 1.75%; when CAPLCM was added to Clab+PAFEM, by comparing No7 and No4, accuracy improved by 2.15%, recall by 1.86%, precision by 2.37%, and $F_1$-score by 1.09%; when CAPLCM is added to Clab+PAFEM+MTT, by comparing No8 and No5, accuracy improves by 3.40%, recall by 5.17%, precision by 0.6%, and $F_1$-score by 2.94%.

**PAFEM effectiveness:** adding PAFE to the base model, accuracy improved by 2.32%, recall by 6.83%, precision by 3.39%, and $F_1$-score by 5.03% by comparing No4 and No1 in Table 6; when PAFE is added to Clab+MTT, by comparing, by comparing No5 and No2 in Table 6, the accuracy is improved by 2.19%, recall by 4.14%, precision by 3.47%, and $F_1$- score by 3.83%; adding PAFE to Clab+CAPLCM+MTT, by comparing No8 and No6 in Table 6, the accuracy is improved by 3.18%, recall by 7.82%, and $F_1$-score by 5.03%. Recall improves by 7.82%, precision reaches by 1.52%, and $F_1$-score obtains by 5.29%.

4. 5. 2 Ablation Results of the segmentation task

For the segmentation task, we evaluate the effect of seven networks, which is shown in Table 7. And experimental results are shown in Table 8.

Table 7 Abbreviations and their interpretation of the models we validate in the ablation experiments of the segmentation task

| No | Abbreviations | Segb | CADCM | Clab | MTT | PAFEM | CAPLCM |
|----|---------------|------|-------|------|-----|-------|--------|
| 1 | Segb | √ | | | | | |
| 2 | SCC | √ | √ | √ | | | |
| 3 | SCCM | √ | √ | √ | √ | | |
| 4 | SCCP | √ | √ | √ | | √ | |
| 5 | SCCPM | √ | √ | √ | √ | √ | |
| 6 | SCCCM | √ | √ | √ | √ | | √ |
| 7 | SCCPCM | √ | √ | √ | √ | √ | √ |

Table 8 Presentation details of different models for segmentation tasks. Bold black text indicates the highest scores of all methods in the corresponding metrics.

| No | Method | Class | $F_1$-score | Precision | Recall | IoU |
|----|--------|-------|-------------|-----------|--------|-----|
| 1 | Segb | | 97.73 | **97.79** | 97.67 | 95.56 |
| 2 | SCC | | 97.76 | 97.72 | 97.81 | 95.62 |
| 3 | SCCM | | 97.76 | 97.72 | 97.80 | 95.62 |
| 4 | SCCP | Tongue | 97.75 | 97.72 | 97.78 | 95.60 |
| 5 | SCCPM | | 97.75 | 97.72 | 97.78 | 95.60 |
| 6 | SCCCM | | 97.76 | 97.72 | 97.81 | 95.62 |
| 7 | SCCCPM | | **97.77** | 97.72 | **97.81** | **95.63** |
| 1 | Segb | | 63.69 | 62.47 | 65.18 | 46.75 |
| 2 | SCC | | 64.50 | 63.94 | 65.29 | 47.62 |
| 3 | SCCM | | 64.43 | 63.81 | 65.29 | 47.55 |
| 4 | SCCP | ToothMarks | 64.21 | 63.37 | 65.29 | 47.31 |
| 5 | SCCPM | | 64.18 | 63.30 | 65.29 | 47.28 |
| 6 | SCCCM | | 64.50 | 63.94 | 65.29 | 47.62 |
| 7 | SCCCPM | | **64.52** | **63.97** | 65.29 | **47.64** |

**CADCM and Clab effectiveness:** when CADCM and Clab are added to the base model, by comparing No2 and No1 in Table 8, the $F_1$-score for the class of the tooth marks improves by 0.81%, Precision by 1.47%, Recall by 0.11%, and IoU by 0.87%; as for the class of tongue the $F_1$-score improves by 0.03%, Recall by 0.14%, and IoU by 0.06%.

**CAPLCM effectiveness:** as shown in Table 8, when CAPLCM is added to SCCM, by comparing No6 with No3, $F_1$-score for the class of tooth marks rises by 0.07%, Precision by 0.13%, and IoU by 0.07%; Recall for the tongue class improves by 0.01%; when CAPLCM is added to SCCPM, by comparing No7 with No5, the $F_1$-score of the class of tooth marks improves by 0.34%, Precision by 0.67%, and IoU by 0.36%; the $F_1$-score of the tongue class improves by 0.02%, Recall by 0.03%, and IoU by 0.03%.

**PAFEM effectiveness:** when PAFEM is added to SCCCM, by comparing No6 and No7 in Table 8, $F_1$-score improves by 0.02%, Precision reaches by 0.03%, and IoU rises by 0.02% for the class of tooth marks, and $F_1$-score improves by 0.01% and IoU by 0.01% for the tongue class.

# 5. Discussion

5. 1 Clinical applicability

The model Ammonia-Net effectively classifies tongues using RGB images captured by standard cameras, covering HT, LT, MT, and ST classes, which is as well trained for segmentation. Compared to other advanced models, Ammonia-Net outperforms SOTA on several metrics. The diagnosis of tooth-marked tongue relies entirely on the domain experts. Training highly skilled specialists in tooth-marked tongue diagnosis over a short time is challenging. In comparison, Ammonia-Net leverages artificial intelligence to swiftly implement diagnostic procedures across various machines. It is cost effective and assist in telemedicine, diagnoses, and medical students in their education. Consequently, Ammonia-Net proves valuable in overcoming the constraints of limited medical resources. Moreover, according to the confusion matrix of the four-level classification task, our model successfully identifies all potential patient cases, serving as a critical safeguard against missed diagnoses in both AI-autonomous and AI-assisted diagnostic processes.

5. 2 Incorrect case analyzation

As shown in Fig. 3, the confusion matrix shows that Ammonia-Net exhibits substantial recognition ability for HT, followed by MT, and relatively weak for LT and ST. This indicates that Ammonia-Net is proficient in identifying tooth-marked and healthy tongues. However, there is still potential for improvement in distinguishing between different tooth-marked tongue classes. In case of incorrect classification, as shown in Fig. 8. Possible explanations for the challenges in the classification of the tooth-marked tongue are as follows: (1) similar features between different classes, which leads to blurred interclass boundaries; (2) the data quality is limited. Captured images are not focused on the tongue, which means the resolution of the tongue area is low resulting in loss of features; (3) data volume is limited and imbalance exists between classes, for example, the HT and MT account for 34% and 27.2% respectively, which may make difficult to learn the features of classes with less data.

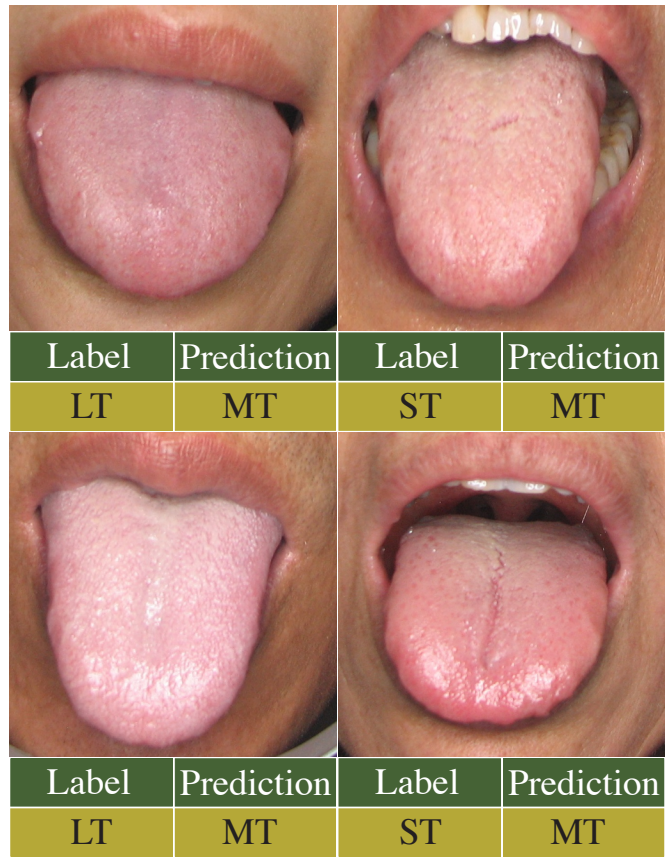

**Fig. 8. Error cases of Ammonia-Net in tooth-marked tongue classification. The ground truth and predicted results are marked below each image.**

For the segmentation task, deviations are observed between the predictions of our model and the ground truth for the pixel. The possible reasons are as follows: (1) some tongue bodies have a covering such as the moss on the tongue. The covering may be uneven as shown in the third row of Fig 9. This makes the features of the dentition naturally noisy. The color of the tooth marks varies a lot between images of TTD-2020. The consistency in features is weak. (2) The color of the tooth mark is close to other parts of the tongue. The boundary between the tooth mark and the tongue is blurred because of no standard illumination and orientation while capturing images as well as physiological problems of the tongue, such as inflammation. In other words, there is a flatter excess at the pixel value level, which makes the segmentation of the tooth marks more challenging task. Fig 8 shows the prediction error of the pixel points.

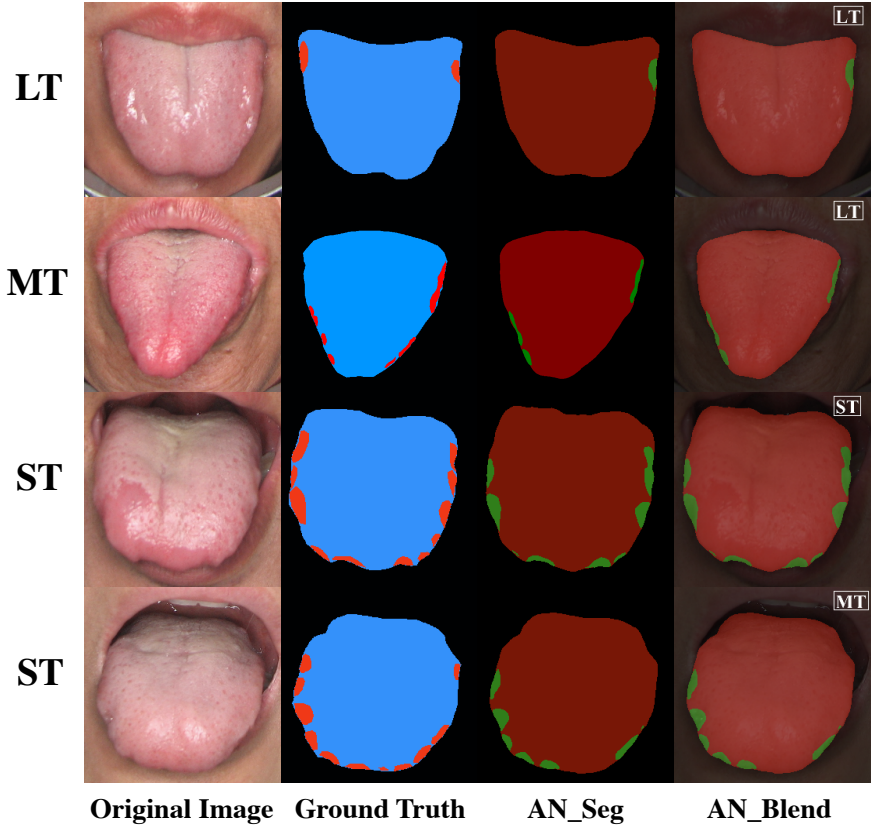

**Fig. 9.** Error predictions given by Ammonia-Net, with the ground truth of each given vertically on the left. The horizontal text at the bottom describes the origin of the images in each column. Where "AN _Seg" indicates the segmentation result given by Ammonia-Net and "AN_Blend" denotes the final output of Ammonia-Net, which is a fusion of the original image and the segmentation output, with the classification result displayed in the top right corner.

## 6. Conclusion

We propose a multi-task joint learning model, namely Ammonia-Net, for the diagnosis of tooth-marked tongue with RGB images. More specifically, we combine the segmentation of tooth marks and tongues with the classification task of the tooth-marked tongue. Multi-task joint training allows the proposed network to accomplish the multiple tasks simultaneously. And to the best of our knowledge, we make the first attempt at using segmentation output of tooth marks to assist in tooth-marked tongue classification. Moreover, the classification output helps to improve the segmentation performance by erasing the tooth marks for healthy tongues. In optimizing the network, we employ a multi-task objective function to further optimize the results of classification. Comprehensive multi-metric experiments conducted on the TTD-2020 dataset, which prove that Ammonia-Net has better results compared to other SOTA methods.

Moving forward, one method to enhance the performance is by leveraging tongue data with distance information. Additionally, integrating prior knowledge about visceral status connected to tooth marks offers a promising approach for directly diagnosing diseases via tongue images.

Acknowledgments

This work is supported by grants from National Natural Science Foundation of China (31970752)；Science, Technology, Innovation Commission of Shenzhen Municipality (JCYJ20190809180003689, JSGG20200225150707332, JCYJ20220530143014032, ZDSYS20200820165400003, WDZC20200820173710001, WDZC20200821150704001, JSGG20191129110812708); Shenzhen Bay Laboratory Open Funding (SZBL2020090501004) and Shandong University startup funding.23